# Structured Evolution with Compact Architectures for Scalable Policy Optimization

Krzysztof Choromanski [* 1]   Mark Rowland [* 2]   Vikas Sindhwani [1]   Richard E. Turner [2]   Adrian Weller [2 3]

## Abstract

We present a new method of blackbox optimization via gradient approximation with the use of structured random orthogonal matrices, providing more accurate estimators than baselines and with provable theoretical guarantees. We show that this algorithm can be successfully applied to learn better quality compact policies than those using standard gradient estimation techniques. The compact policies we learn have several advantages over unstructured ones, including faster training algorithms and faster inference. These benefits are important when the policy is deployed on real hardware with limited resources. Further, compact policies provide more scalable architectures for derivative-free optimization (DFO) in high-dimensional spaces. We show that most robotics tasks from the OpenAI Gym can be solved using neural networks with less than 300 parameters, with almost linear time complexity of the inference phase, with up to 13x fewer parameters relative to the Evolution Strategies (ES) algorithm introduced by Salimans et al. (2017). We do not need heuristics such as fitness shaping to learn good quality policies, resulting in a simple and theoretically motivated training mechanism.

## 1. Introduction

The goal of reinforcement learning (RL) is to find, through trial and error, a feedback policy that prescribes how an agent should optimally act in a dynamic, uncertain environment. If a family of policies mapping states to actions is parameterized by $\theta \in \mathbb{R}^d$ – say $d$ weights of a deep neural network $\pi_\theta$ – optimizing the long-term objectives of the agent via direct policy search can be cast as a maximization problem of the form,

$$\max_{\theta \in \mathbb{R}^d} F(\theta), \quad (1)$$

where the objective $F : \mathbb{R}^d \to \mathbb{R}$ measures the expected total reward of the policy $\pi_\theta$. One would then expect to iteratively improve the policy via gradient ascent,

$$\theta_{k+1} = \theta_k + \eta \nabla F(\theta_k).$$

However, when the environment is stochastic and implemented either in blackbox physics simulators or as opaque wrappers around a real mechanical system such as a robot, the objective function $F$ is only accessible via noisy and expensive function evaluations, with no gradients $\nabla F(\theta_k)$ available. In such situations, one can turn to derivative-free optimization (DFO) (Conn et al., 2009).

Given that DFO techniques usually require $O(d)$ times more iterations than standard gradient methods (Nesterov & Spokoiny, 2017; Jamieson et al., 2012), conventional wisdom dictates that it is unreasonable to expect them to work well for high-dimensional problems; indeed Conn et al. (2009) remark that "the scale of the problems that can currently be efficiently solved by derivative-free methods is still relatively small and does not exceed a few hundred variables even in easy cases". It is therefore no surprise that pure blackbox methods were largely abandoned in favor of "greybox" policy gradient methods (Schulman et al., 2017; 2015; Lillicrap et al., 2015) that exploit the Markovian structure of the RL setting with considerable success. However, in recent work, Salimans et al. (2017) demonstrated that a very simple randomized finite difference DFO approach, derived as an "evolutionary strategy" (ES), is comparable to state of the art policy gradient schemes for training deep neural network controllers on a variety of simulated robotics tasks in MuJoCo and on Atari games. These results, together with the relative simplicity, generality and parallelizability of DFO methods has generated renewed interest in them for policy optimization.

In this paper, our motivation is to significantly improve the overall efficiency of DFO-based optimization of neural network policies, noting that Salimans et al. (2017) required considerable computational resources – a distributed ES implementation running on thousands of workers – to get competitive results. We propose two complementary strategies that are shown to be highly effective in a comprehensive set of experiments:

- *Structured Exploration*: We demonstrate theoretically and empirically that random orthogonal and Quasi Monte Carlo (QMC) finite difference directions are much more effective for parameter exploration than random Gaussian directions used in (Salimans et al., 2017). We also outline a fast, discrete construction using rows of randomized Hadamard matrices. Our experiments include a detailed comparison of various DFO schemes on a collection of 212 benchmark optimization problems from (Moré & Wild, 2009) and 12 continuous control tasks from the OpenAI Gym benchmark suite.

- *Compact policies*: By imposing a parameter sharing structure on the policy architecture, we are able to dramatically reduce the problem dimensionality without losing accuracy. We show that on MuJoCo tasks in the OpenAI Gym benchmark suite, most problems can be solved with networks up to 13 times smaller than those of Salimans et al. (2017) – barely 300 parameters each – without losing any accuracy. The weight matrices in these networks are Toeplitz, supporting fast inference using fast Fourier transforms. We expect these results to be of independent interest for resource constrained (e.g. low power, limited storage) settings arising pervasively in mobile robotics and other embedded applications.

*Equal contribution   [1]Google Brain Robotics   [2]University of Cambridge   [3]The Alan Turing Institute. Correspondence to: Krzysztof Choromanski <kchoro@google.com>, Mark Rowland <mr504@cam.ac.uk>.

## 2. Gaussian Smoothings and Monte Carlo Gradient Estimation

We start by introducing the notion of *Gaussian smoothing* (Nesterov & Spokoiny, 2017; Staines & Barber, 2012) of the objective function $F$ in Equation (1),

$$
\begin{aligned}
F_\sigma(\theta) &= \mathbb{E}_{\varepsilon \sim \mathcal{N}(\mathbf{0},I)}\left[F(\theta + \sigma\varepsilon)\right] \\
&= (2\pi)^{-d/2} \int_{\mathbb{R}^d} F(\theta + \sigma\varepsilon) e^{-\frac{1}{2}\|\varepsilon\|_2^2} \mathrm{d}\varepsilon,
\end{aligned}
\qquad (2)
$$

where $\sigma > 0$ plays the role of a smoothing parameter. $F_\sigma$ is obtained by perturbing $F$ at a given point along Gaussian directions and averaging the evaluations. Informally, $F_\sigma$ is "nicer" than $F$ in the sense that it is differentiable everywhere even if $F$ is not, and any initial smoothness properties of $F$ carry over to $F_\sigma$; for precise statements, see (Nesterov & Spokoiny, 2017). The Euclidean distance between gradients of $F$ and $F_\sigma$ can be pointwise-bounded (see Lemma 3 in (Nesterov & Spokoiny, 2017)) so that it is plausible to replace the problem in Equation (1) with

$$
\max_{\theta \in \mathbb{R}^d} F_\sigma(\theta), \qquad (3)
$$

while expecting a high quality solution to the original problem.

The optimal value of Problem (3) lower-bounds that of Problem (1). One may hope that optimal parameters $\theta^*$ for Problem (3) are close to those of Problem (1), though in general no guarantees can be given. Indeed, the loss surface of $F_\sigma$ differs qualitatively from that of $F$, in a sense leading to flatter solutions, which can yield better robustness of the solution (Hochreiter & Schmidhuber, 1997; Lehman et al., 2017).

### 2.1. Estimating Gradients of Gaussian Smoothings

The gradient of $F_\sigma$ is given by

$$
\begin{aligned}
\nabla F_\sigma(\theta) &= \frac{1}{\sigma}\mathbb{E}_{\varepsilon \sim \mathcal{N}(\mathbf{0},I)}\left[F(\theta + \sigma\varepsilon)\varepsilon\right] \\
&= (2\pi)^{-d/2} \int_{\mathbb{R}^d} F(\theta + \sigma\varepsilon) e^{-\frac{1}{2}\|\varepsilon\|_2^2} \varepsilon \mathrm{d}\varepsilon.
\end{aligned}
\qquad (4)
$$

In practice, this gradient is intractable, and must be estimated. The gradient $\nabla F_\sigma(\theta)$ can be estimated via a variety of Monte Carlo estimators; in this paper, we investigate variance reduction techniques for three such estimators.

**Vanilla ES gradient estimator.** The first estimator is derived via the Monte Carlo REINFORCE (Williams, 1992) (or score function) estimator, and coincides with a standard Monte Carlo estimator of the expectation appearing in Equation (4):

$$
\widehat{\nabla}_N^{\mathrm{V}} F_\sigma(\theta) = \frac{1}{N\sigma} \sum_{i=1}^N F(\theta + \sigma\varepsilon_i)\varepsilon_i, \qquad (5)
$$

where $(\varepsilon_i)_{i=1}^N \stackrel{\mathrm{iid}}{\sim} \mathcal{N}(\mathbf{0}, I)$ can be interpreted as parameter exploration directions – that is, perturbations in parameter space to be explored. We refer to this gradient estimator as the *vanilla ES gradient estimator*.

**Antithetic ES gradient estimator.** Secondly, we consider a version of the vanilla ES gradient estimator augmented with antithetic variables, as in (Salimans et al., 2017), given by

$$
\widehat{\nabla}_N^{\mathrm{AT}} F_\sigma(\theta) = \frac{1}{2N\sigma} \sum_{i=1}^N (F(\theta + \sigma\varepsilon_i)\varepsilon_i - F(\theta - \sigma\varepsilon_i)\varepsilon_i), \qquad (6)
$$

where again $(\varepsilon_i)_{i=1}^N \stackrel{\mathrm{iid}}{\sim} \mathcal{N}(\mathbf{0}, I)$. The two terms appearing in the summand are contributed by a sample $\varepsilon_i \sim \mathcal{N}(\mathbf{0}, I)$ and its antithetic counterpart $-\varepsilon_i$. We refer to this gradient estimator as the *antithetic ES gradient estimator*, owing to its use of antithetic Monte Carlo samples.

**Forward finite-difference ES gradient estimator.** Finally, other unbiased estimators of $\nabla F_\sigma(\theta)$ can be obtained by introducing a control variate into the vanilla ES gradient estimator. In particular, the *forward finite-difference ES gradient estimator* of $\nabla F_\sigma(\theta)$ is defined as follows:

$$\widehat{\nabla}_N^{\text{FD}} F_\sigma(\theta) = \frac{1}{N\sigma} \sum_{i=1}^{N} (F(\theta + \sigma\varepsilon_i) - F(\theta))\varepsilon_i, \quad (7)$$

where again $(\varepsilon_i)_{i=1}^N \stackrel{\text{iid}}{\sim} \mathcal{N}(\mathbf{0}, I)$. The forward finite-difference (FD) ES estimator is natural to consider as it renders the vanilla ES gradient estimator in (5) invariant to shifts of $F$ by a constant, without incurring the cost of an additional function evaluation, as in the antithetic ES estimator.

Natural questions arise as to the statistical and computational efficiencies of these estimators, and whether one dominates the other. In fact, neither forward FD nor antithetic dominates the other, as we shall soon show in Examples 2.1 and 2.2. We first define an error measure.

For a given estimator $\widehat{\nabla} F_\sigma(\theta)$ of the gradient $\nabla F_\sigma(\theta)$ evaluated at point $\theta$, the *mean squared error* (MSE) of $\widehat{\nabla} F_\sigma(\theta)$ is defined as: $\text{MSE}(\widehat{\nabla} F_\sigma(\theta)) = \mathbb{E}[\|\widehat{\nabla} F_\sigma(\theta) - \nabla F_\sigma(\theta)\|_2^2]$, i.e. the expected squared distance.

**Example 2.1** (Forward FD outperforms antithetic)**.** Let $F(\mathbf{x}) = \langle \mathbf{a}, \mathbf{x} \rangle$, for some fixed $\mathbf{a} \in \mathbb{R}^d$, so that in this case, $F_\sigma(\theta) = \mathbb{E}_{\phi \sim \mathcal{N}(\theta, \sigma^2 I)}[F(\phi)] = \langle \mathbf{a}, \theta \rangle$, and $\nabla F_\sigma(\mathbf{0}) = \mathbf{a}$. Note that the estimators $\widehat{\nabla}_2^{\text{AT}} F_\sigma(\mathbf{0})$ and $\widehat{\nabla}_3^{\text{FD}} F_\sigma(\mathbf{0})$ both require $4$ evaluations of $F$ to compute. A straightforward calculation reveals that $\text{MSE}(\widehat{\nabla}_3^{\text{FD}} F_\sigma(\mathbf{0})) < \text{MSE}(\widehat{\nabla}_2^{\text{AT}} F_\sigma(\mathbf{0}))$.

**Example 2.2** (Antithetic outperforms forward FD)**.** Let $F(\mathbf{x}) = \|\mathbf{x}\|^2$, so that $F_\sigma(\theta) = \mathbb{E}_{\phi \sim \mathcal{N}(\theta, \sigma^2 I)}[F(\phi)]$. Then $\nabla F_\sigma(\mathbf{0}) = \mathbf{0}$, and we have $\widehat{\nabla}_1^{\text{AT}} F_\sigma(\mathbf{0}) = \mathbf{0}$ almost surely, achieving zero MSE, whilst $\widehat{\nabla}_1^{\text{FD}} F_\sigma(\mathbf{0}) \neq \mathbf{0}$ almost surely. Note that both estimators require two evaluations of $F$.

To contrast these estimators against those that we introduce next, we will also refer to them as *iid estimators*, to emphasize that their exploration directions are sampled independently.

## 3. Variance Reduction via Orthogonality and Quasi-Monte Carlo Exploration

We will now introduce new strategies for improving the quality of the gradient estimators introduced above through judicious choices of the joint distribution over exploration directions $(\varepsilon_i)_{i=1}^N$.

Without loss of generality we will focus on the Antithetic ES estimator $\widehat{\nabla}_N^{\text{AT}} F_\sigma(\theta)$, Eqn. 6; the constructions for the other two estimators are completely analogous.

### 3.1. Gaussian Orthogonal Exploration

We enforce orthogonality conditions on the Gaussian perturbations for parameter exploration. The corresponding estimator is given as follows.

**Definition 3.1.** The *orthogonal centered FD estimator of the Gaussian smoothing antithetic ES gradient estimator* is given by

$$\widehat{\nabla}_N^{\text{AT,ort}} F_\sigma(\theta) = \frac{1}{2N\sigma} \sum_{i=1}^N (F(\theta + \sigma\varepsilon_i')\varepsilon_i' - F(\theta - \sigma\varepsilon_i')\varepsilon_i') \quad (8)$$

where the $(\varepsilon_i')_{i=1}^N$ are all marginally distributed as $\mathcal{N}(\mathbf{0}, I)$, and the joint distribution is defined as follows: if $N \leq d$, then the vectors are conditioned to be orthogonal almost-surely. If $N > d$, then each consecutive set of $d$ vectors is conditioned to be orthogonal almost-surely, with distinct sets of $d$ vectors remaining independent.

The above exploration scheme for $N \leq d$ can be encoded by a structured random matrix, where by "structured" we mean a random matrix for which the rows have a non-trivial dependence structure, obtained by stacking together exploration directions $\epsilon_i'$. We term this construction a *Gaussian orthogonal matrix*, owing to the fact that each row is marginally Gaussian, but all rows are mutually orthogonal almost-surely, and denote it by $\mathbf{G}_{\text{ort}}$.

Our next result (proof in the Appendix) shows that by imposing different directions for gradient estimation to be exactly orthogonal (rather than just having zero expected dot-product), we obtain an estimator characterized by strictly lower MSE than the corresponding iid gradient estimator. This result motivates the use of orthogonal directions for gradient estimation in the blackbox optimization setting.

**Theorem 3.2.** The orthogonal antithetic ES gradient estimator $\nabla_N^{\text{AT,ort}} F_\sigma(\theta)$ is unbiased, and yields lower MSE than the antithetic ES gradient estimator $\widehat{\nabla}_N^{\text{AT}} F_\sigma(\theta)$. For $N \leq d$, the improvement in MSE is quantified by:

$$\text{MSE}(\widehat{\nabla}_N^{\text{AT,ort}} F_\sigma(\theta)) = \text{MSE}(\widehat{\nabla}_N^{\text{AT}} F_\sigma(\theta)) - \frac{N-1}{N} \|\nabla F_\sigma(\theta)\|_2^2.$$

**Remark 3.3.** The conclusion of Theorem 3.2 holds even if evaluations of the function $F$ are each corrupted by independent mean-zero noise, as may be the case in Monte Carlo roll-outs in a reinforcement learning context.

### 3.2. Discrete Orthogonal Exploration

We have seen in Theorem 3.2 that statistical improvements in gradient estimators can be achieved by using orthogonal exploration directions, but we have not yet commented on the computational aspects of this method. In this regard, gradient estimators relying on Gaussian orthogonal directions have one disadvantage – they require Gram-Schmidt orthogonalization of an unstructured Gaussian matrix at every iteration of the optimization procedure, in order to obtain a set of orthogonal exploration directions. As we will see in Sections 5 and 6, it is still reasonable to perform this orthogonalisation in the setting of compact neural network policies, where the number of parameters is few hundred and thus Gram-Schmidt orthogonalization is performed on relatively small matrices. We will show that compact neural networks with such quantities of parameters suffice to learn good quality policies for most policy optimization settings considered.

However, even for higher-dimensional regimes, one can still take advantage of orthogonal exploration directions without incurring the high orthogonalization cost (as before, without loss of generality we assume that the number of exploration directions $N$ satisfies $N \leq d$, where $d$ stands for parameter dimensionality). To achieve this, we replace the collection of pairwise orthogonal vectors $\{\epsilon'_1, ..., \epsilon'_N\}$ with marginal Gaussian distribution with a collection $\{\mathbf{d}_1, ..., \mathbf{d}_N\}$ of perturbation directions, where $\mathbf{d}_i^\top$ is the $i^{th}$ row of a structured discrete random matrix $\mathbf{M}_{\text{struct}} \in \mathbb{R}^{N \times d}$. In other words, we replace $\mathbf{G}_{\text{ort}}$ with a structured random matrix that can be constructed without any orthogonalization preprocessing, but has similar properties to $\mathbf{G}_{\text{ort}}$. Next we give examples of such matrices $\mathbf{M}_{\text{struct}}$.

**Gradient estimation via random Hadamard-Rademacher matrices.** Here we take the structured random matrix $\mathbf{M}_{\text{struct}}$ to be of the form $\mathbf{M}_{\text{struct}}^{\text{HAD}} = d^{-\frac{k-1}{2}} \mathbf{H}\mathbf{D}_1 \mathbf{H}\mathbf{D}_2 \cdot ... \cdot \mathbf{H}\mathbf{D}_k$, where $\mathbf{D}_i$s stand for independent copies of a random diagonal matrix with diagonal entries given by independent Rademacher random variables (that is, with distribution $\text{Unif}(\{-1, +1\})$) and $\mathbf{H}$ is a Kronecker-product Hadamard matrix of the form $\mathbf{H}_1^{\otimes l}$, where

$$\mathbf{H}_1 = \begin{pmatrix} -1 & 1 \\ 1 & 1 \end{pmatrix},$$

and $\mathbf{H}_1^{\otimes l}$ stands for the Kronecker product of $l$ copies of $\mathbf{H}_1$ for some $l \in \mathbb{N}_+$.

Hadamard matrix-vector products can be computed in subquadratic time, due to the Fast Hadamard Transform, and construction of random diagonal matrices $\mathbf{D}_i$ at every iteration of the optimization procedure has complexity linear in the dimensionality of the parameter space. Note also that since rows of $\mathbf{H}$ are orthogonal, this is also the case for the matrix $d^{-\frac{k-1}{2}} \mathbf{H}\mathbf{D}_1 \mathbf{H}\mathbf{D}_2 \cdot ... \cdot \mathbf{H}\mathbf{D}_k$. Furthermore, it is easy to check that the squared $L_2$-norms of the rows of $d^{-\frac{k-1}{2}} \mathbf{H}\mathbf{D}_1 \mathbf{H}\mathbf{D}_2 \cdot ... \cdot \mathbf{H}\mathbf{D}_k$ are the same as the expected squared lengths of the rows of $\mathbf{G}_{\text{ort}}$. Thus $\mathbf{M}_{\text{struct}}^{\text{HAD}}$ defined in this way resembles $\mathbf{G}_{\text{ort}}$, but does not require any preprocessing for the orthogonalization. Such constructions have been studied in other contexts where a fast approximation to uniform Haar measure on the group of orthogonal matrices is required, such as in dimensionality reduction (Choromanski et al., 2017), locality-sensitive hashing (Andoni et al., 2015), and approximate kernel methods (Yu et al., 2016).

Note that the matrices $\mathbf{M}_{\text{struct}}^{\text{HAD}}$ defined above are of sizes which are powers of two. If the input vectors do not have this property, a standard 0-padding mechanism can be applied to embed them in the desired higher-dimensional space (for further details, see e.g. (Yu et al., 2016)).

**Orthogonal gradient estimators via length renormalization.** One can obtain other variants of structured matrices $\mathbf{M}_{\text{struct}}$ by renormalizing the rows of $\mathbf{G}_{\text{ort}}$ and $\mathbf{M}_{\text{struct}}^{\text{HAD}}$. In particular, the matrix $\mathbf{G}_{\text{ort}}^{\text{renorm}}$ is obtained from $\mathbf{G}_{\text{ort}}$ by sampling their lengths from the distribution of rows from $\mathbf{M}_{\text{struct}}^{\text{HAD}}$ and vice versa, matrix $\mathbf{M}_{\text{struct}}^{\text{HAD,renorm}}$ is obtained from $\mathbf{M}_{\text{struct}}^{\text{HAD}}$ by sampling their lengths from the distribution of the rows of $\mathbf{G}_{\text{ort}}$. In all four variants the rows of the structured matrix constitute a random orthogonal basis and furthermore, the expected squared lengths of the rows are the same (two of them are in fact deterministic).

### 3.3. Quasi-Monte Carlo Exploration

In addition to our structured exploration methods, we also propose the use of Quasi-Monte Carlo (QMC) approximations (Caflisch, 1998; Dick & Pillichshammer, 2010; Avron et al., 2016) to the integral in Equation (4) to construct a gradient estimator. To the best our knowledge, QMC approximations have not been explored in this context. QMC techniques constitute a family of methods for numerical integration, in which an integration problem over the unit cube is approximated using a deterministic point set $S$ as follows,

$$\int_{[0,1]^d} f(x)dx \approx \frac{1}{|S|} \sum_{w \in S} f(w).$$

In such QMC approximations, the point set $S$ is a *low-discrepancy sequence* that offers faster rates of convergence than a random point set as used in Monte-Carlo integration. Informally, "discrepancy" is a measure of non-uniformity of the point set, and several constructions of low-discrepancy sequences are available in the literature (Dick & Pillichshammer, 2010). In particular, we used generalized Halton sequences (see e.g. (Dick & Pillichshammer, 2010) which admit fast construction and have performed well on closely related tasks (Avron et al., 2016). We apply QMC to the integral in Equation (5) using a standard transformation into an integration problem over the unit cube; see (Avron et al., 2016) for similar calculations.

## 4. Learning Compact Policies

Deep neural network policies are typically composed of non-linear vector-valued transforms of the form, $f(\mathbf{x}, \mathbf{M}) = s(\mathbf{M}\mathbf{x})$, where $s$ is an elementwise nonlinearity, $x$ is an input vector, and $\mathbf{M}$ is an $m \times n$ matrix of parameters. When $\mathbf{M}$ is a large general dense matrix, the cost of storing $mn$ parameters and computing matrix-vector products in $O(mn)$ time can make it prohibitive to deploy such policies on mobile agents with specialized resource-constrained (i.e., low power and storage) onboard hardware. Learning compact policies is not only important for inference in such settings, but is particularly appealing for training: the smaller the policy search space, the more robust ES/DFO can be expected to be. A number of network compression schemes have been proposed in the literature (Han et al., 2015; Sainath et al., 2013; Sindhwani et al., 2015). In this paper, we experiment with parameter sharing schemes, such as imposing a Toeplitz structure on the weight matrices. Recall that a Toeplitz matrix $\mathbf{T} \in \mathbb{R}^{k \times l}$ satisfies the property that $\mathbf{T}_{ij}$ depends only on $i - j$.

A $n \times n$ Toeplitz matrix has constant diagonals and supports fast matrix-vector products via Fast Fourier Transforms. The family of Toeplitz matrices can be vastly generalized to low-displacement rank matrices for more complex capacity-accuracy-time tradeoffs. Our experiments show that on 12 benchmark MuJoCo RL tasks, Toeplitz matrices offer $13\times$ compression relative to the networks used in (Salimans et al., 2017) with superior DFO training curves.

## 5. Distributed Implementation

The Monte Carlo estimators of the gradient of the Gaussian smoothing of Section 2.1 enable distributed implementation. In the setting with $L$ machines, the exploration directions $\{\epsilon_1, ..., \epsilon_N\}$ or $\{\epsilon'_1, ..., \epsilon'_N\}$ can be partitioned among all the workers. Each worker computes only the part of the sum from Equation (6) (or other equations related to other Monte Carlo estimators and direction choices) corresponding to directions assigned to it. Averaging over the $L$ workers is performed by one central worker (or may be conducted via peer-to-peer architectures). This provides a scalable mechanism capable of handling thousands of directions, as needed for larger policy networks (Salimans et al., 2017).

### 5.1. Further Features of Structured Approaches

We discuss the impact of the orthogonal variance reduction schemes introduced in Section 3 on the efficacy of structured evolution strategies in a distributed framework. Specifically, we consider using $k$ Hadamard-Rademacher $\mathbf{HD}$ blocks.

**Communication Complexity.** Structured evolution strategies also achieve a low communication overhead. To start, the workers and master agree on an initial random seed and a fixed assignment of matrix row numbers to the workers. At each iteration, each worker generates the same $k$ Hadamard-Rademacher $\mathbf{HD}$ matrices, computes the product if $k > 1$, and selects its appropriate row. No further information about perturbations need be exchanged – just the results of function evaluations by the workers which are collected by the master.

**Computational Complexity.** As noted in Section 3, constructing a set of uniformly random orthogonal exploration directions requires additional linear algebra operations to be performed in comparison to workers using iid exploration directions. Further, this work must be performed locally on each worker if increased communication costs are to be avoided. However, as noted in Section 3.2, high-quality discrete approximations to uniformly random orthogonal matrices can be used, at a fraction of the computational cost. More precisely, in comparison to the $\mathcal{O}(d)$ cost associated with reconstructing an iid perturbation from a random seed, perturbations based on $k > 1$ random Hadamard-Rademacher blocks may be reconstructed in $\mathcal{O}(kd \log d)$ time per worker per perturbation direction, via the Fast Hadamard Transform.

Specifically, to compute the $i^{th}$ row of the matrix $\mathbf{HD}_1 \cdot ... \cdot \mathbf{HD}_k$, the worker must compute $(\mathbf{HD}_1 \cdot ... \cdot \mathbf{HD}_k)^\top \mathbf{e}_i$, where $\mathbf{e}_i$ is a $\{0, 1\}$-vector with only the $i^{th}$ dimension nonzero. This product can be computed in time $\mathcal{O}(kd \log(d))$ via the Fast Walsh-Hadamard Transform. For the special case $k = 1$ the cost is linear as in the unstructured case, since each worker can store the rows of a fixed Hadamard matrix $\mathbf{H}$ that correspond to the directions it is interested in. The computation of the randomized version of that row defining a perturbation direction with a matrix $\mathbf{D}_1$ can be conducted in linear time. If perturbation directions are defined by the rows of the Gaussian orthogonal matrix $\mathbf{G}_{\text{ort}}$, then each worker must first construct a random Gaussian matrix and then Gram-Schmidt orthogonalization to obtain $\mathbf{G}_{\text{ort}}$, which can be done in time $\mathcal{O}(d^3)$ per worker.

On the topic of computational complexity, we reiterate that the fast linear-algebraic methods that can be used to perform inference with the compactly-parameterized networks described in Section 4 lead to a reduction in computational cost for each worker, in comparison to the use of unstructured networks. Indeed, in contrast to standard quantization mechanisms, where compression is performed on a fully-parameterized trained network, these architectures enable us to train on the already compressed network. This provides much more scalable training methods. The structured neural network architectures we propose also provide faster inference than unstructured baselines (log-linear vs. quadratic time complexity) which may be important in practice.

# 6. Experiments

We consider two main experimental settings. Firstly, we compare structured evolution strategies against iid baseline approaches on a collection of 212 lower-dimensional blackbox optimization tasks drawn from a benchmark suite developed by the DFO community (Moré & Wild, 2009), where the estimated gradients are used to perform optimization. Secondly, in the context of RL we train neural network policies on 12 Mujoco continuous control tasks from the OpenAI Gym collection. Here, we compare total rewards collected when structured exploration directions are used in conjunction with compact architectures, against the baseline method of (Salimans et al., 2017) where full networks are used with unstructured random Gaussian exploration directions.

## 6.1. Lower-Dimensional Blackbox Optimization with Structured Evolution Strategies

We exhibit the methods described above on the derivative-free optimization benchmarking suite (Moré & Wild, 2009), consisting of 53 low-dimensional blackbox optimization tasks. There are four variants of these tasks so that the total number of benchmark problems is 212:

- smooth, in which the objective functions are smooth;
- nondiff, in which the objective functions are non-differentiable;
- noisy3, in which the smooth functions are perturbed by deterministic noise; and
- wild3 in which the smooth functions are perturbed by stochastic noise.

We compare antithetic ES estimators of the following varieties: IID, which selects each exploration direction independently, as in Equation (6); ORT, which selects exploration directions to be almost-surely orthogonal (but marginally uniform), as in Equation (8); HD, which selects exploration directions according to the fast Hadamard-Rademacher random matrices described in Section 3.2 (we use $k = 1$ matrix blocks in all cases); and QMC, which selects directions jointly according to a Quasi-Monte Carlo strategy, specifically using a generalized Halton sequence (see e.g. (Dick & Pillichshammer, 2010) for further details, and Section 11 of the Appendix for precise parameter settings). For the optimization of the functions, we use MATLAB's inbuilt `fminunc` gradient-based optimization routine (running the BFGS Quasi-Newton method) in combination with the gradients estimated by the ES methods; full details are given in Section 11 of the Appendix.

We compare the quality of final objective value in each optimization task, and the number of function evaluations before optimization terminated across the methods. To quantitatively summarize the performance of the methods in these two domains across the wide range of optimization tasks at hand, we compute a "normalized score", where the best method for a particular task receives a score of 0, the worst method receives a score of 1, and the remaining methods receive scores that linearly interpolate between these two extremes, based on their raw performance – see Figure 1. We also compute a ranking (best to worst) of the four methods on each optimization task (a rank of 1 corresponding to best performance, and 4 worst), and average these across the 53 optimization tasks – see Figure 4 in the Appendix. We observe that in all cases, structured exploration outperforms IID exploration.

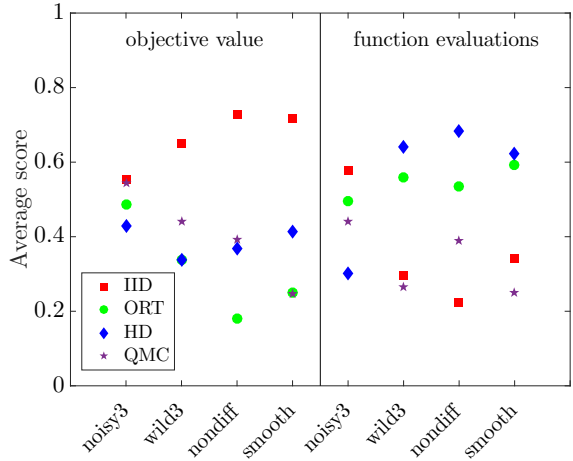

*Figure 1.* Average score across DFO tasks for the antithetic estimator (6) with a variety of exploration distributions. The standard deviation of the exploration distribution was $10^{-6}$ for all methods. Scores based on final objective value are given on the left-hand side of the figure, whilst score based on function evaluations are given on the right-hand side. Lower scores are better.

## 6.2. Learning Structured Policies via Structured Gradient Estimation

**Reinforcement learning environments.** We consider the a collection of reinforcement learning OpenAI Gym (Brockman et al., 2016) tasks, summarized in Table 1.

**Monte Carlo gradient estimators.** We test all three variants of the Monte Carlo gradient estimator discussed in the paper, namely: antithetic ES, forward finite-difference ES and vanilla ES. For each environment and architecture under consideration we choose the variant that corresponded to training with the highest reward. We observed that for different environments, different Monte Carlo variants were optimal, which supports our remarks in Section 2.1 that there does not exist one universal optimal control variate term.

|  | CP:ST($\mathbf{G}_{\text{ort}}$) | CP:ST($\mathbf{H}$) | FP:UN |
|---|---|---|---|
| Swimmer | 253 | 253 | 1408 |
| Ant | 362 | 254 | 4896 |
| HalfCheetah | 266 | 254 | 2174 |
| Hopper | 257 | 254 | 1536 |
| Humanoid | 636 | 510 | 13664 |
| Walker2d | 266 | 254 | 1824 |
| Pusher | 273 | 255 | 2048 |
| Reacher | 256 | 256 | 1189 |
| Striker | 273 | 255 | 2048 |
| Thrower | 273 | 255 | 2048 |
| ContMountCar | 246 | 246 | 1184 |
| Pendulum | 247 | 247 | 1216 |

Table 1. OpenAI Gym RL tasks benchmarked and number of parameters in neural network policy architectures for the following settings: CP:ST($\mathbf{G}_{\text{ort}}$) using compact policies and structured directions from $\mathbf{G}_{\text{ort}}$, CP:ST($\mathbf{H}$) using compact policies and and structured directions from $\mathbf{M}_{\text{struct}}^{\text{HAD}}$ and baseline FP:UN using unstructured "full" policies and unstructured directions.

**Architectures.** We consider here the following training methods for learning RL policies $\pi_\theta$:

- **FP:UN**: Full feedforward Policy neural networks, with UNstructured directions.
- **CP:UN**: Compact Policy neural networks, with UNstructured directions.
- **CP:ST**: Compact Policy neural networks, with STructured directions.

The FP:UN neural network architectures consist of input layer (state), two hidden layers of size 32 each and one output layer (proposed action). We use $\tanh$ nonlinearities. The structured neural networks coupled with Gaussian orthogonal matrices $\mathbf{G}_{\text{ort}}$ or matrices $\mathbf{M}_{\text{struct}}^{\text{HAD}}$ (with $k = 1$ since for $k > 1$ similar learning plots were obtained) for gradient estimation use two hidden layers and Toeplitz matrices encoding connection weights between consecutive layers as well as $\tanh$ nonlinearities. With Gaussian orthogonal exploration directions, each hidden layer was of size $h = 41$; with Hadamard mechanisms the sizes were chosen in such a way that the total number of parameters was close to (but not exceeding) **256** for smaller environments and close to but not exceeding **512** for larger ones (Humanoid), see Table 1.

**Optimization algorithm.** The optimization was conducted with the use of AdamOptimizer and the same fixed learning rate $\alpha$ that was used in (Salimans et al., 2017). We did not use any heuristics such as fitness shaping.

| ES timesteps ($\times 10^6$) | CP:ST | (Salimans et al., 2017) |
|---|---|---|
| Swimmer | 3 | 1.39 |
| HalfCheetah | 0.6 | 2.88 |
| Hopper | 30 | 31.6 |
| Walker2d | 8 | 37.9 |

Table 2. Number of ES timesteps per worker for our CP:ST mechanism for policy learning, compared to the ES algorithm of Salimans et al. (2017). It is not clear but appears likely that we use significantly fewer total workers.

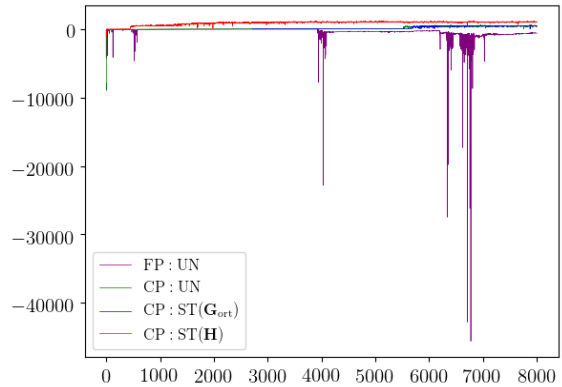

Figure 2. Learning curves for the Ant environment

**Resources.** We use TensorFlow distributed synchronous infrastructure with at most 400 workers (1 cpu / worker). In Table 2 we present the number of ES timesteps per worker used by our mechanism CP:ST for four of our twelve tasks that are also reported in (Salimans et al., 2017), and compare to the total number of ES timesteps per worker for the ES algorithm as reported in (Salimans et al., 2017). The exact number of workers used in the experiments described in (Salimans et al., 2017) is not clear, but it appears likely that we use significantly fewer in our method, due to the compact parameterization of our policies.

**Results - quality of the structured policies with structured gradient estimation.** We observed that on most OpenAI Gym tasks considered, it was not possible to learn competitive policies with the FP:UN mechanism (note that in contrast to (Salimans et al., 2017) we did not apply any additional heuristics such as fitness shaping), whereas CP:ST learnt high quality policies for most tasks. In particular, we managed to solve the following environments with CP:ST: Swimmer, Ant, HalfCheetah, Hopper, Walker2d, Pusher, Reacher, Striker, Continuous Mountain Car, Pendulum.

We also observed that CP:ST outperformed CP:UN on most tasks, confirming that structured exploration with random orthogonal directions leads to higher rewards than the baseline unstructured one. The mechanism CP:ST was superior to CP:UN on the following tasks: Continuous Mountain Car, HalfCheetah, Hopper, Pendulum, Reacher, Swimmer (both: Gaussian orthogonal and Hadamard), Ant (Hadamard), Humanoid and Walker2d (Gaussian orthogonal). Fully unstructured neural network architectures turned out to be better than Toeplitz structured neural networks on just one out of twelve OpenAI Gym tasks (see: Fig 4).

|     | CP:ST($\mathbf{G}_{\text{ort}}$) | CP:ST($\mathbf{H}$) | CP:UN | FP:UN |
| --- | --- | --- | --- | --- |
| AN  | 509.2 | **1144.57** | 566.42 | -12.78 |
| SW  | 370.36 | **370.53** | 368.38 | 150.90 |
| HC  | **3619.92** | 3273.94 | 1948.26 | 2660.76 |
| HO  | 99883.29 | **99969.47** | 99464.74 | 1540.10 |
| HU  | **1842.7** | 84.13 | 1425.85 | 509.56 |
| WA  | **9998.12** | 9974.60 | 9756.91 | 456.51 |
| PU  | -48.07 | -43.04 | **-36.71** | -46.91 |
| RE  | **-4.29** | -10.31 | -73.04 | -145.39 |
| ST  | -112.78 | -87.27 | **-52.56** | -63.35 |
| TH  | -349.72 | -243.88 | -265.60 | **-192.76** |
| CMC | 92.05 | **93.06** | 90.69 | -0.09 |
| PE  | -128.48 | **-127.06** | -3278.60 | -5026.27 |

*Table 3.* Total rewards obtained on different robotics OpenAI Gym tasks for different neural network architectures and exploration strategies. Highest rewards are bold (AN:Ant, SW:Swimmer, HC:HalfCheetah, HO:Hopper, HU:Humanoid, WA:Walker2d, PU:Pusher, RE:Reacher, ST:Striker, TH:Thrower, CMC:Continuous Mountain Car, PE:Pendulum).

Learning curves for all OpenAI Gym tasks are presented in Figures 9, 10, 11, 12, in Section 12 of the Appendix. We reproduce the plots for the Ant environment in Figure 2, to illustrate the form they take in general.

Each plot consists of two curves: total reward as a function of iteration (total reward) as well as a curve that tracks the max reward upto the current iteration (max total reward). The symbol $\mathbf{H}$ in these plots stands for structured gradient estimation with directions encoded by matrices $\mathbf{M}_{\text{struct}}^{\text{HAD}}$. For the Ant environment we substantially reduced the number of steps per roll-out to $s = 500$, in the learning phase, but then tested the best policy across all iteration steps using full roll-outs. The mechanism CP:ST with matrices $\mathbf{G}_{\text{ort}}$ provided the highest reward ($R = 9249$) and the second best mechanism was CP:UN (with reward $R = 8615$).

In Table 4 we summarized learning curves from training phase by recording the total reward obtained by an optimal policy found by each training method on each of the twelve tested environments from OpenAI Gym.

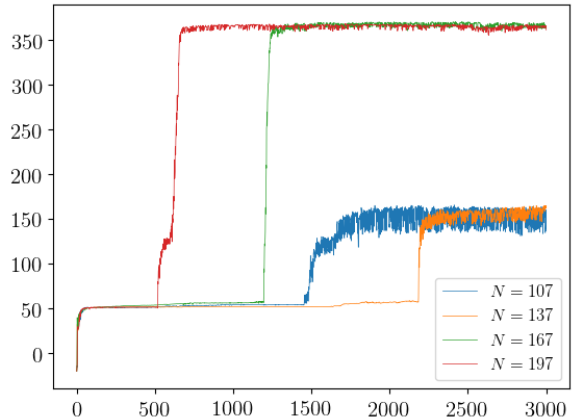

*Figure 3.* Learning curves for Swimmer env with different number of Gaussian orthogonal directions ($N$) used for structured exploration in the CP:ST setup with $d = 253$ parameters.

**Results - reducing number of structured directions for gradient estimation.** We also conducted experiments showing how the quality of the learned policy depends on the number of structured directions chosen to estimate the gradient. Our results show that not only do structured architectures enable to reduce the number of roll-outs of the environment (which is a bottleneck of all the computations), but further reduction can be achieved by using $N < d$ structured directions ($d$ stands for the number of parameters of the neural network), since structured orthogonal directions are evidently of superior quality than unstructured. In contrast, in (Salimans et al., 2017) $N = d$ unstructured directions are applied. We managed to learn the optimal policy of the Swimmer using only $N = \mathbf{167}$ roll-outs per iteration of the optimization procedure for the structured policy with Gaussian orthogonal directions for exploration (having $d = 253$ parameters), see: Figure 3). In comparison, Salimans et al. (2017) required $N = \mathbf{4864}$ roll-outs per iteration.

## 7. Conclusion

In this paper, we have proposed two complementary methods for derivative-free optimization in the context of reinforcement learning: structured evolution strategies, and compact policy networks. We showed that they can be successfully applied to provide scalable blackbox optimization algorithms and used in reinforcement learning to learn good quality policies. Natural questions for future work are to what extent other matrix structures can be exploited to achieve compact policy networks, and whether other variance-reduction methods are available for use in evolution strategies.

## Acknowledgements

We thank Matthias Bauer, Wessel Bruinsma and María Lomelí for helpful comments, as well as the anonymous reviewers. MR acknowledges support by the UK Engineering and Physical Sciences Research Council (EPSRC) grant EP/L016516/1 for the University of Cambridge Centre for Doctoral Training, the Cambridge Centre for Analysis. RET is supported by Google as well as EPSRC grants EP/M0269571 and EP/L000776/1. AW acknowledges support from the David MacKay Newton research fellowship at Darwin College, The Alan Turing Institute under EPSRC grant EP/N510129/1 & TU/B/000074, and the Leverhulme Trust via the CFI.

# Appendix

## 8. Proofs

### 8.1. Proof of Theorem 3.2

*Proof.* Unbiasedness follows immediately since the $(\varepsilon'_i)_{i=1}^N$ each have the same marginal distribution as the $(\varepsilon_i)_{i=1}^N$. For the MSE claim, we provide a proof for the case $N \leq d$; the general case is analogous. Let $(\varepsilon_i)_{i=1}^N$ be iid $N(0, I)$, and let $(\varepsilon'_i)_{i=1}^N$ be marginally $\mathcal{N}(0, I)$ and almost-surely orthogonal. Write

$$F^{(i)} = \frac{1}{2\sigma} \left( F(\theta + \sigma \varepsilon_i)\varepsilon_i - F(\theta - \sigma \varepsilon_i)\varepsilon_i \right),$$

$$F'^{(i)} = \frac{1}{2\sigma} \left( F(\theta + \sigma \varepsilon'_i)\varepsilon'_i - F(\theta - \sigma \varepsilon'_i)\varepsilon'_i \right),$$

for each $i = 1, \ldots, N$, and note that

$$\begin{aligned}
\mathrm{MSE}(\widehat{\nabla}_N^{\mathrm{AT,ort}} F_\sigma(\theta)) &= \mathbb{E}\left[\left\| \frac{1}{N} \sum_{i=1}^N F'^{(i)} - \nabla F_\sigma(\theta) \right\|_2^2 \right] \\
&= \mathbb{E}\left[ \left\| \frac{1}{N} \sum_{i=1}^N F'^{(i)} \right\|_2^2 \right] - \|\nabla F_\sigma(\theta)\|_2^2 \\
&= \frac{1}{N^2} \left( \sum_{i=1}^N \mathbb{E}\left[\|F'^{(i)}\|_2^2\right] + \sum_{i \neq j} \mathbb{E}\left[\langle F'^{(i)}, F'^{(j)} \rangle\right] \right) - \|\nabla F_\sigma(\theta)\|_2^2. \quad (9)
\end{aligned}$$

By analogous reasoning, we have the following expression for the MSE of the iid estimator:

$$\mathrm{MSE}(\widehat{\nabla}_N^{\mathrm{AT}} F_\sigma(\theta)) = \frac{1}{N^2} \left( \sum_{i=1}^N \mathbb{E}\left[\|F^{(i)}\|_2^2\right] + \sum_{i \neq j} \mathbb{E}\left[\langle F^{(i)}, F^{(j)} \rangle\right] \right) - \|\nabla F_\sigma(\theta)\|_2^2. \quad (10)$$

Since $F^{(i)}$ and $F'^{(i)}$ are equal in distribution, we have $\mathbb{E}\left[\|F^{(i)}\|_2^2\right] = \mathbb{E}\left[\|F'^{(i)}\|_2^2\right]$. Now note that $\langle F'^{(i)}, F'^{(j)} \rangle = 0$ almost surely for $i \neq j$, so $\mathbb{E}\left[\langle F'^{(i)}, F'^{(j)} \rangle\right] = 0$ in Equation (9). Note also that since $F^{(i)}$ and $F^{(j)}$ are independent for $i \neq j$, we have $\mathbb{E}\left[\langle F^{(i)}, F^{(j)} \rangle\right] = \langle \nabla F_\sigma(\theta), \nabla F_\sigma(\theta) \rangle = \|\nabla F_\sigma(\theta)\|_2^2 \geq 0$ in Equation (10). Therefore, the stated result follows. $\square$

## 9. Implementation details

In this section, we give further information on the construction of exploration directions using Hadamard-Rademacher random matrices and quasi-Monte Carlo strategies, as well as precise details of the Toeplitz parametrisations used for policy networks in the experiments.

### 9.1. Exploration directions with Hadamard-Rademacher random matrices and quasi-Monte Carlo

Here, we provide precise algorithmic details as to how Hadamard-Rademacher random matrices and quasi-Monte Carlo sequences can be used to construct exploration directions, complementing the discussion in Sections 3.2 and 3.3.

Algorithm 1 sets out the computation required to generate exploration directions from Hadamard-Rademacher random matrices.

Algorithm 2 describes the computation required to generate exploration directions from a quasi-Monte Carlo sequence. The first step of the algorithm is to draw a set of samples which resemble draws from $\mathrm{Unif}([0, 1]^d)$. Rather than sampling i.i.d. from this distribution, instead a call to a standard QMC sampler is used – these samples are designed to "fill the space" more efficiently that i.i.d. samples from $\mathrm{Unif}([0, 1]^d)$ typically would. There are many QMC sampling algorithms that may be used; in our experiments, we use generalized Halton sequences (additional details are given in Section 11.1), but see (Dick & Pillichshammer, 2010) for a extensive survey of commonly-used QMC sampling methods. The second step is to transform these samples from occupying the unit hypercube $[0, 1]^d$, to approximating a collection of multivariate Gaussian samples in $\mathbb{R}^d$; this is acheived by applying the Gaussian CDF coordinate-wise to the hypercube samples.

---

**Algorithm 1** Hadamard-Rademacher exploration directions

1: Sample the matrices $\mathbf{D}_1, \ldots, \mathbf{D}_k$ by drawing i.i.d. Rademacher random variables for each diagonal entry.
2: Set $\mathbf{G} = \mathbf{I}$, the identity matrix
3: **for** j=1,…,k **do**
4:     Set $\mathbf{G} \leftarrow \mathbf{D}\mathbf{G}$
5:     Compute $\mathbf{G} \leftarrow \mathbf{H}\mathbf{G}$ via the Fast Walsh-Hadamard Transform.
6: **end for**
7: Compute $\mathbf{G} \leftarrow d^{-\frac{k-1}{2}}\mathbf{G}$
8: Use the resulting rows of $\mathbf{G}$ as exploration direction, in place of Gaussian vectors $(\varepsilon_i)_{i=1}^N$ in the estimators described in Section 2.

---

**Algorithm 2** Quasi-Monte Carlo exploration directions

1: Generate a sequence of quasi-Monte Carlo samples $(\mathbf{x}_i)_{i=1}^N \subset [0,1]^d$ via a standard QMC algorithm.
2: For each sample $\mathbf{x}_i$ ($i = 1, \ldots, N$), compute the *transformed sample* $\varepsilon_i$ given by applying the standard Normal CDF to each coordinate of $\mathbf{x}_i$.
3: Use the $(\varepsilon_i)_{i=1}^N$ as exploration directions in the estimators described in Section 2.

---

### 9.2. Toeplitz network structures

The Toeplitz structure is enforced by encoding this part of the network with vectors of size: $m + n - 1$, where $m$ stands for the number of rows and $n$ for the number of columns. The entire network is vectorized and in the inference phase de-vectorized into a sequence of structured matrices. Note that we never explicitly backpropagate through the network, we only run forward passes. To update parameters of the network, we always use vectorized representations.

## 10. Related work

In this section, we briefly mention other work related to our approach. Whilst our methods are focused on variance reduction for *isotropic* Gaussian smoothings of an objective function $F$, there has been much work on adapting the smoothing online, to reflect the local properties of $F$ at the current set of parameters; a principal example of such an approach is CMA-ES (Hansen et al., 2003). These adaptive approaches have been shown to yield considerable improvements in performance versus isotropic baselines in certain circumstances. Here, we observe that these adaptive approaches are complementary to our variance reduction techniques, and in principle these variance reduction techniques could be extended to methods that invoke covariance adaptation (by, for example, enforcing that exploration directions are orthogonal under a whitening transform with respect to the current covariance matrix). We leave it as an open question for future as to how such exploration methods can be implemented in a computational efficient manner across a distributed system. These ES approaches differ from other recent methods for continuous control, such as DDPG (Lillicrap et al., 2015), in that they do not take advantage of any Markov structure in the environment. We remark, however, that exploration in DDPG is achieved by injecting Gaussian noise into the actions of an agent, and there may be interesting further work in understanding whether the variance reduction techniques studied here are applicable in these contexts too.

## 11. Experimental Details for Section 6.1

### 11.1. Further Experimental Details

We provide full details of the experimental setup for the optimisation problems solved in Section 6.1. Gradient estimates from each ES strategy were supplied to MATLAB's built-in `fminunc` gradient-based optimisation function, using the `quasi-newton` option. The final objective value reported for a given optimisation problem and exploration method was given by the output of the `fminunc` method, and the number of function evaluations reported for a given optimsiation problem and exploration method was the total number of function evaluations recorded during the call to `fminunc`. The number of exploration directions was taken to be equal to the dimensionality of the optimisation problem in all circumstances, unless otherwise stated.

For the QMC method described in the main paper, we MATLAB's built-in `haltonset` function to generate a generalized Halton sequence in the unit hypercube, apply a reverse-radix scrambling, and then apply coordinate-wise inverse Gaussian cumulative density functions to obtain multivariate Gaussian samples. The `leap` and `skip` parameters of `haltonset` were set to 700 and 1000 respectively. The deterministic reverse-radix scrambling is applied to the quasi-Monte Carlo stream to the stream of points via MATLAB's built-in `scramble` function.

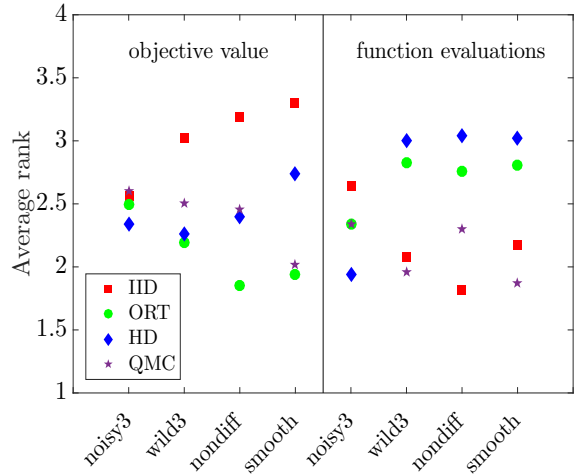

*Figure 4.* Average rankings across DFO tasks for the antithetic estimator (6) with a variety of exploration distributions. The standard deviation of the exploration distribution was $10^{-6}$ for all methods. Rankings based on final objective value are given on the left-hand side of the figure, whilst rankings based on function evaluations are given on the right-hand side. Lower ranks are better.

### 11.2. Ranking Comparison

Here, we give the comparison of the methods described in Section 6.1 based on rankings, as described in the main paper. The results are broadly in line with the comparison based on normalized scores.

### 11.3. Further Experiment Results: Varying Exploration Noise

In this section, we study the effect of varying the exploration noise parameter $\sigma$ on the findings of Section 6.1. In Section 6.1, $\sigma$ was set to $10^{-6}$ in all experiments. Here, we give corresponding results with $\sigma$ set to $10^{-7}$ (see Figures 5 and 6) and $10^{-5}$ (see Figures 7 and 8). Overall, the relative behaviour of the exploration methods remains similar as the exploration noise is varied.

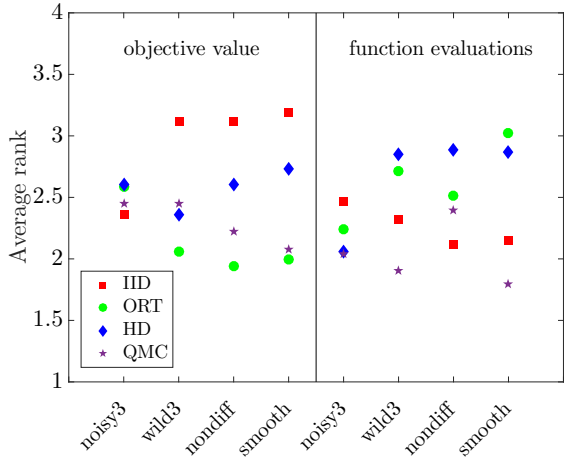

Figure 5. $\sigma = 10^{-7}$, average ranks.

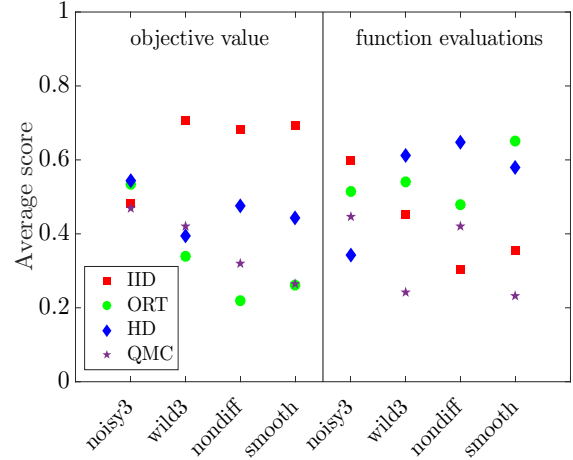

Figure 6. $\sigma = 10^{-7}$, average scores.

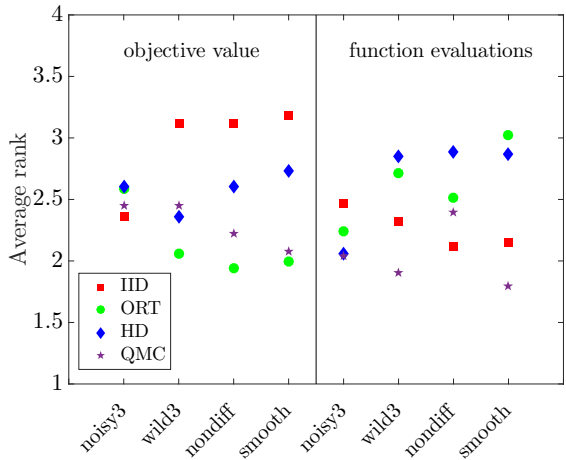

Figure 7. $\sigma = 10^{-5}$, average ranks.

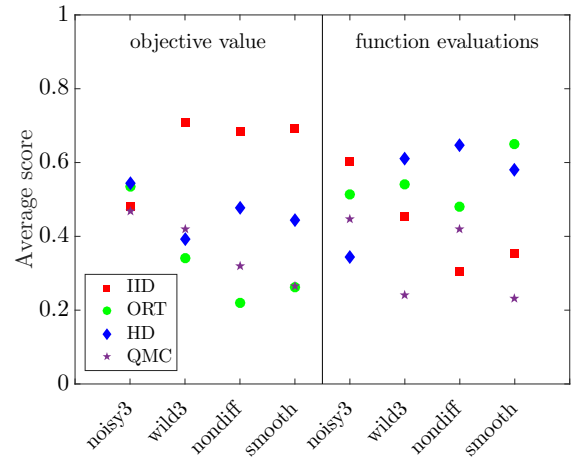

Figure 8. $\sigma = 10^{-5}$, average scores.

## 12. Additional OpenAI Gym Learning Curves

In this section, we provide learning curves for all environments and algorithms described in Section 6.2. We also run experiments on more (20 random seeds), computing the mean reward and standard deviation as well as and add comparison to a standard finite difference method. For the standard finite difference method (FD) all runs are the same since the environment and exploration is completely deterministic, thus standard deviation is 0. The termination of the training procedure is dictated by how much the total reward changed over a specific time interval (if the changes are sufficiently small the optimization procedure terminates).

# Structured Evolution with Compact Architectures for Scalable Policy Optimization

|     | CP:ST($\mathbf{G}_{ort}$) | CP:ST($\mathbf{H}$) | CP:UN | FP:UN | CP:FD | FP:FD |
|-----|---------------------------|---------------------|-------|-------|-------|-------|
| AN  | 514.7/0.2                 | **1150.23**/2.6     | 563.78/0.3 | -13.11/1.2 | 505.00 | -10.23 |
| SW  | 370.0/0.1                 | **371.0**/0.1       | 367.1/1.2 | 151.1/5.2 | 313.22 | 174.48 |
| HC  | **3623.56**/2.3           | 3277.11/3.3         | 1942.55/1.4 | 2656.39/2.7 | 1672.65 | 3011.22 |
| HO  | 99888.11/4.2              | **99893.21**/2.7    | 99460.36/1.8 | 1536.00/2.2 | 94032.65 | 10032.69 |
| HU  | **1849.3**/2.2            | 88.13/1.9           | 1430.01/2.8 | 511.93/5.2 | 1325.79 | 624.78 |
| WA  | **10001.05**/2.8          | 9980.63/3.2         | 9754.48/3.2 | 459.33/4.2 | 9003.11 | 531.20 |
| PU  | -49.68/0.2                | -43.33/0.3          | **-35.25**/0.1 | -47.37/0.3 | -49.57 | -51.42 |
| RE  | **-4.11**/0.24            | -12.31/0.44         | -74.31/0.67 | -149.63/1.2 | -85.62 | -181.57 |
| ST  | -113.62/1.3               | -88.77/0.3          | **-49.43**/2.3 | -66.61/0.2 | -51.06 | -90.94 |
| TH  | -361.55/5.2               | -241.55/1.1         | -267.32/3.4 | **-190.51**/1.8 | -415.49 | -382.12 |
| CMC | 91.89/0.2                 | **94.11**/0.1       | 90.03/0.4 | -0.11/0.1 | 90.79 | -0.11 |
| PE  | -128.55/0.1               | **-125.38**/0.82    | -3290.22/5.4 | -5088.34/4.3 | -3582.62 | -6627.83 |

*Table 4.* Mean total rewards obtained from 20 random seeds on different robotics OpenAI Gym tasks and corresponding standard deviations (mean/std) for different neural network architectures and exploration strategies. Additional columns: CP:FD corresponds to the structured neural network and standard finite difference method for gradient approximation; and FP:FD corresponds to the unstructured neural network with standard finite difference method for gradient approximation. For the FD method all runs are the same (see: comment in the main text) thus standard deviation is 0 and we do not report it. Highest rewards are shown in bold (AN:Ant, SW:Swimmer, HC:HalfCheetah, HO:Hopper, HU:Humanoid, WA:Walker2d, PU:Pusher, RE:Reacher, ST:Striker, TH:Thrower, CMC:Continuous Mountain Car, PE:Pendulum).

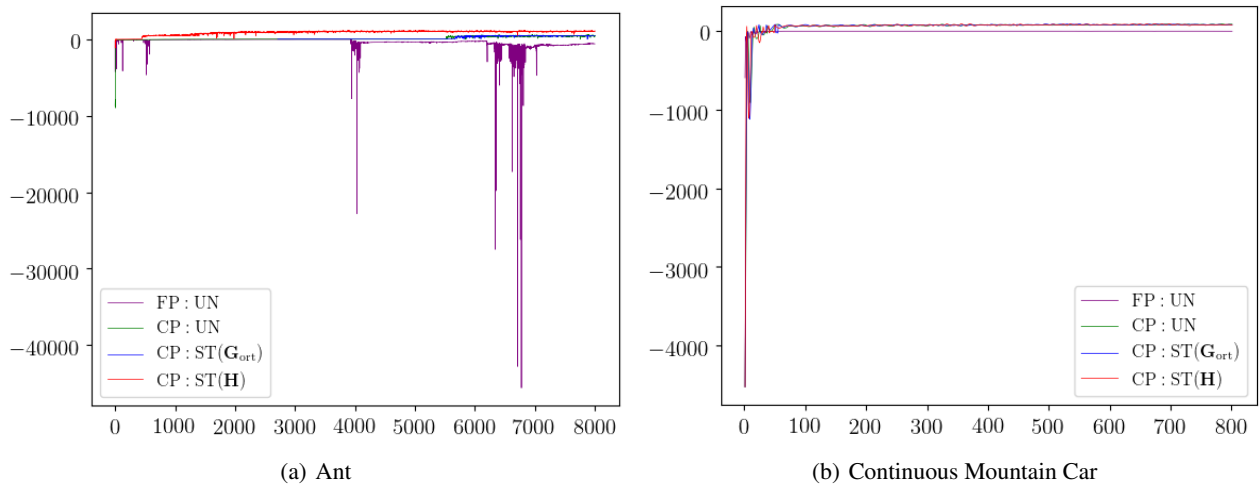

*Figure 9.* Learning curves for different OpenAI Gym envs.

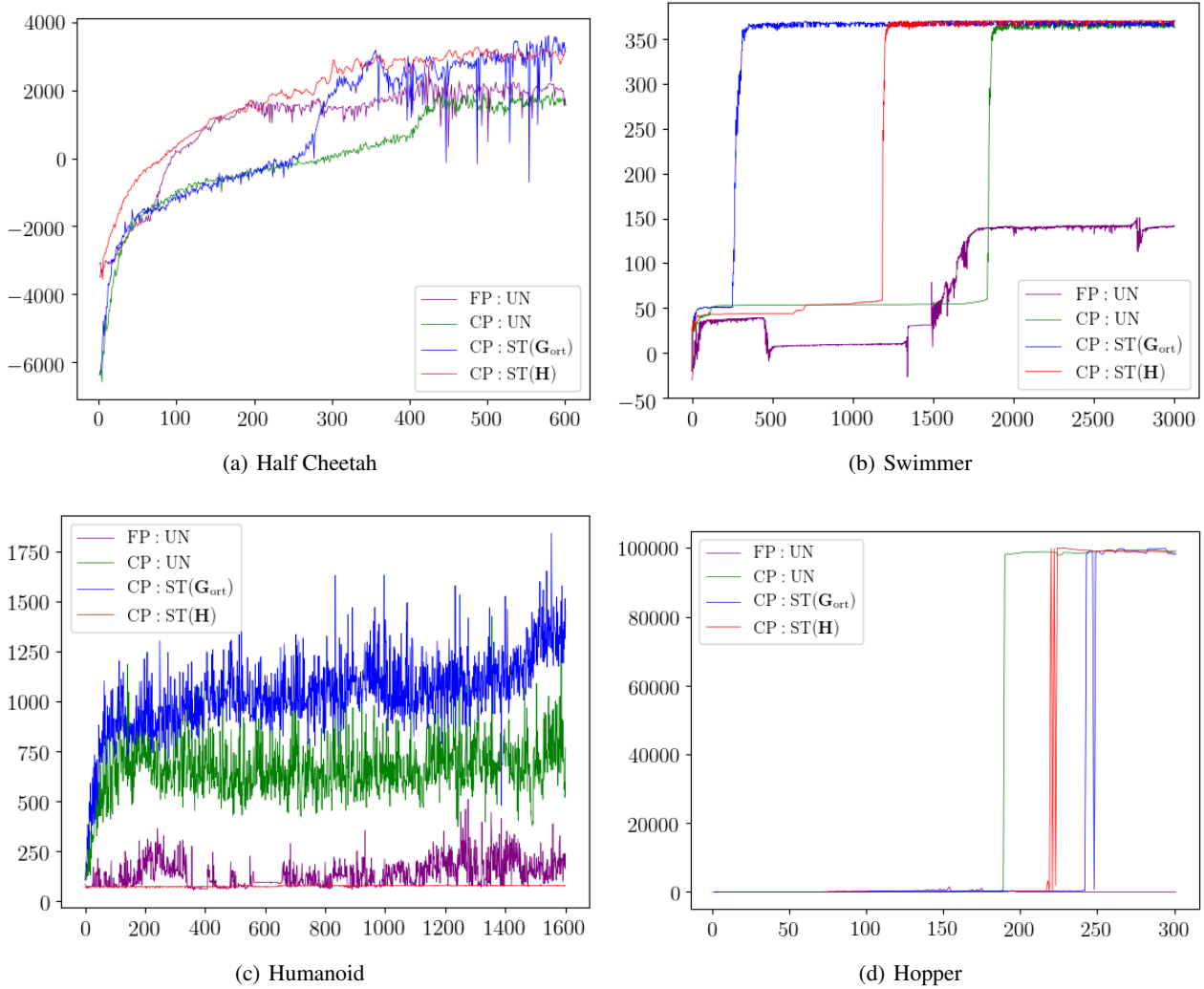

Figure 10. Learning curves for different OpenAI Gym envs.

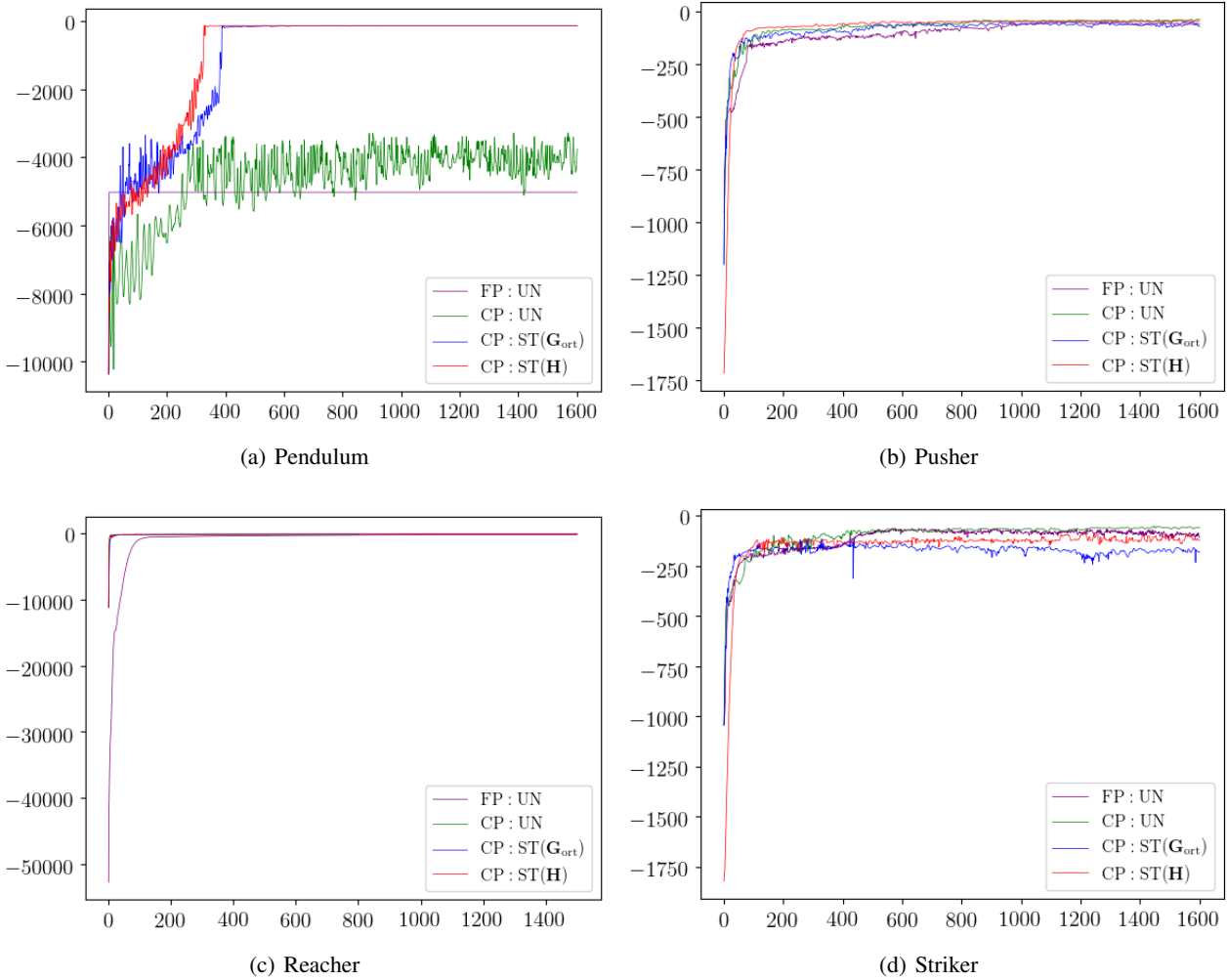

Figure 11. Learning curves for different OpenAI Gym envs.

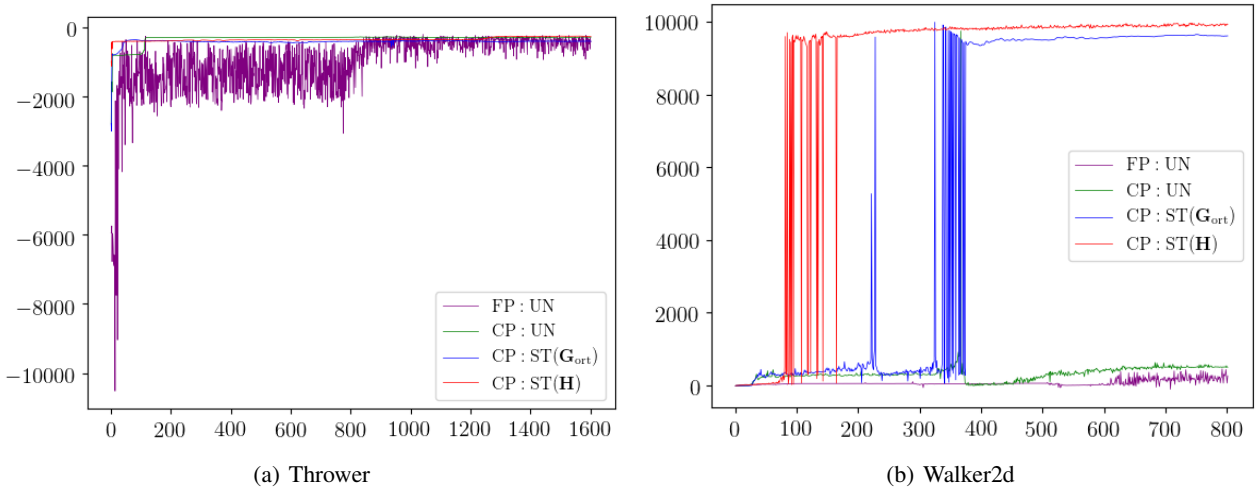

Figure 12. Learning curves for different OpenAI Gym envs.